\documentclass{article}
\usepackage[main, preprint]{arcee}

\usepackage[utf8]{inputenc}
\usepackage[T1]{fontenc}
\usepackage{hyperref}
\usepackage{url}
\usepackage{booktabs}
\usepackage{amsfonts}
\usepackage{bbm}
\usepackage{amsmath}
\usepackage{nicefrac}
\usepackage{microtype}
\usepackage{xcolor}
\usepackage{graphicx}
\usepackage{booktabs}
\usepackage{float}
\usepackage{multicol}

\title{Arcee Trinity Large Technical Report}

\author{%
  Varun Singh$^1$ \\
  \And
  Lucas Krauss$^1$ \\
  \And
  Sami Jaghouar$^2$ \\
  \And
  Matej Sirovatka$^2$ \\
  \And
  Charles Goddard$^1$ \\
  \And
  Fares Obeid$^2$ \\
  \And
  Jack Min Ong$^2$ \\
  \And
  Jannik Straube$^2$ \\
  \And
  Fern$^1$ \\
  \And
  Aria Harley$^1$ \\
  \And
  Conner Stewart$^1$ \\
  \And
  Colin Kealty$^1$ \\
  \And
  Maziyar Panahi$^1$ \\
  \And
  Simon Kirsten$^2$ \\
  \And
  Anushka Deshpande$^1$ \\
  \And
  Anneketh Vij$^1$ \\
  \And
  Arthur Bresnu$^1$ \\
  \And
  Pranav Veldurthi$^1$ \\
  \And
  Raghav Ravishankar$^1$ \\
  \And
  Hardik Bishnoi$^1$ \\
  \And
  DatologyAI Team$^{3,}$\thanks{See appendix for full list of contributors.} \\
  \And
  Mark McQuade$^1$ \\
  \And
  Johannes Hagemann$^2$ \\
  \And
  Lucas Atkins$^1$ \\
}

\begin{document}
\maketitle
\begin{abstract}
We present the technical report for Arcee Trinity Large, a sparse Mixture-of-Experts model with 400B total parameters and 13B activated per token. Additionally, we report on Trinity Nano and Trinity Mini, with Trinity Nano having 6B total parameters with 1B activated per token, Trinity Mini having 26B total parameters with 3B activated per token. The models' modern architecture includes interleaved local and global attention, gated attention, depth-scaled sandwich norm, and sigmoid routing for Mixture-of-Experts. For Trinity Large, we also introduce a new MoE load balancing strategy titled Soft-clamped Momentum Expert Bias Updates (SMEBU). We train the models using the Muon optimizer. All three models completed training with zero loss spikes. Trinity Nano and Trinity Mini were pre-trained on 10 trillion tokens, and Trinity Large was pre-trained on 17 trillion tokens. The model checkpoints are available at \url{https://huggingface.co/arcee-ai}.
\end{abstract}

\begin{figure}[htbp]
  \centering
  \includegraphics[width=0.8\textwidth, trim=2cm 0.3cm 1cm 0.5cm]{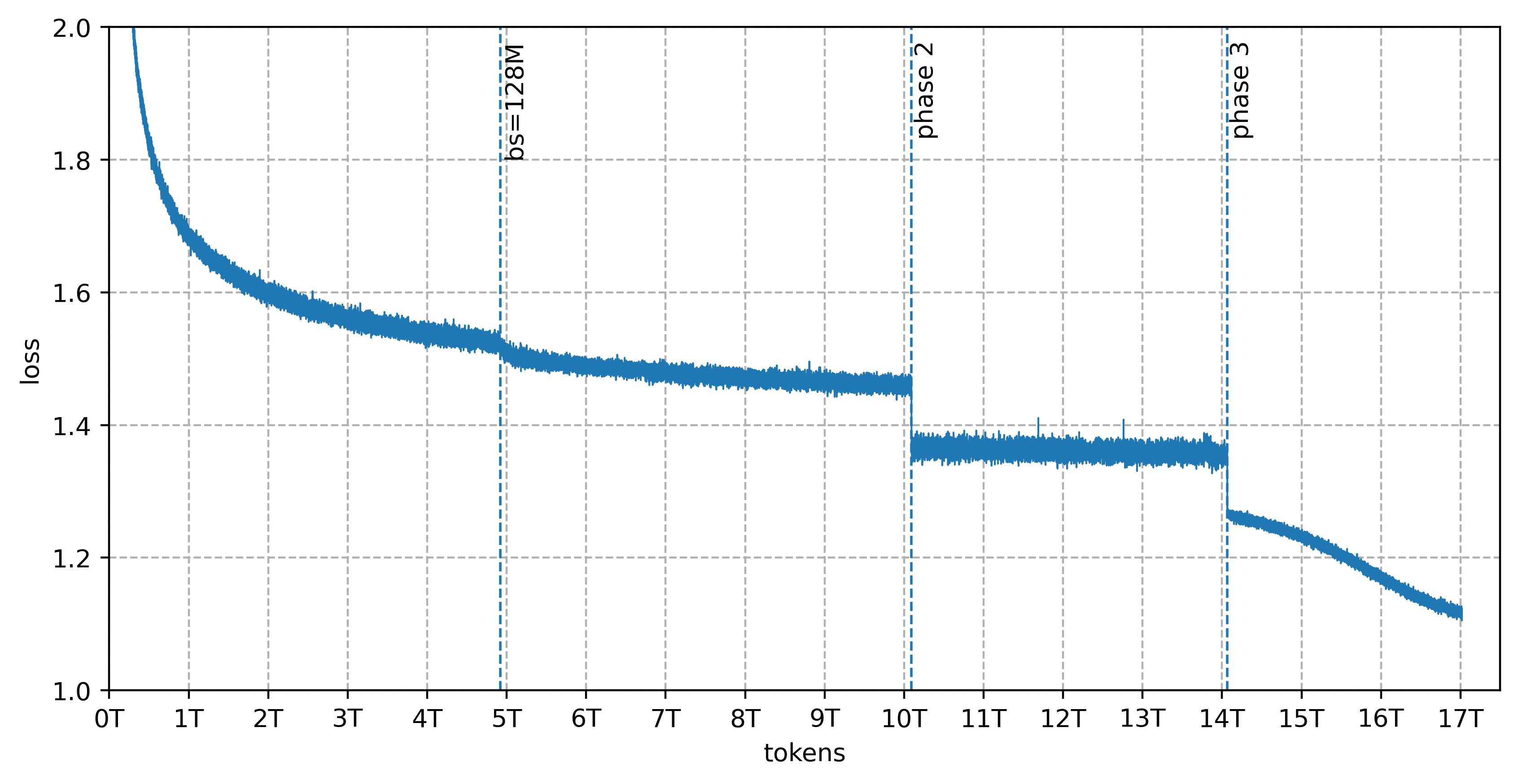}
  \caption{The training loss graph for Trinity Large, with no sub-sampling or smoothing. For clarity, we indicate where the batch size was increased to 128M (134,217,728) tokens, as well as the points where we switch data mixtures.}
  \label{fig:loss}
\end{figure}

\newpage

\section{Introduction}
Large language models (LLMs) are rapidly evolving from use in broad-coverage chatbot applications to general-purpose components for software systems, with deployments spanning document understanding and generation, retrieval-augmented knowledge work, and workflows in which they serve as long-running agents. Tool execution abilities to interact with external environments, code understanding and writing, and long-context management have proven to be key capabilities for LLMs to be deployed effectively. Beyond capabilities, there is also a growing need for inference-time efficiency as workflows and contexts over which the LLMs are required to operate on grow larger and larger. Sparse Mixture-of-Experts (MoE) \citep{shazeer2017outrageouslylargeneuralnetworks} models have emerged as a prominent way for companies to scale up their largest models while being much more efficient and economical to train \citep{deepseekai2025deepseekv3technicalreport, 5team2025glm45agenticreasoningcoding, kimiteam2025kimik2openagentic, coreteam2026mimov2flashtechnicalreport}. In addition to this, recent models incorporate modified attention module designs such as linear attention variants \citep{kimiteam2025kimilinearexpressiveefficient, minimax2025minimax01scalingfoundationmodels, nvidia2025nemotron3nanoopen} or sparse attention variants \citep{deepseekai2025deepseekv32pushingfrontieropen, minicpmteam2025minicpm4ultraefficientllmsend, zhang2026efficientcontextscalinglongcat}, aiming to scale context and throughput without prohibitively increasing training or inference cost. Additionally, the rise of reasoning models \citep{openai2024openaio1card, Guo_2025} has led to inference-time compute scaling as a new paradigm for scaling model performance, enabling models to think via intermediate output for tens to hundreds of thousands of tokens before returning a final answer. These long output chains, often coupled with large inputs that the models need to reason over, further underscore the need for models to be fast and efficient at inference time.

At the same time, the requirements of real-world adoption extend beyond raw benchmark performance. Deployment settings commonly require organizational and regulatory considerations, motivating models that can be audited, hosted, and adapted within environments completely owned by the organizations. In particular, enterprise deployments frequently require clarity about data provenance, licensing, and jurisdictional controls, and therefore benefit from open-weight foundations that can be owned and operated without reliance on opaque third-party checkpoints.

This report presents Trinity Large, the largest member of the Trinity family of open-weight Mixture-of-Experts (MoE) language models. The Trinity family also includes Trinity Nano and Trinity Mini, which were developed as smaller form factors to provide immediately usable open models as well as serve as steps in our scaling ladder that validate the data pipeline, architecture, training recipe, and infrastructure required for Trinity Large.

The Trinity family's architecture is strongly informed from downstream use-cases, primarily the need for efficient training and inference. We adopt an extremely sparse Mixture-of-Experts layer and interleaved local and global attention \citep{yang2025ropenopeagainnew} for efficient inference, and additionally adopt gated attention for stronger context comprehension capabilities. Additionally, the models were trained with the Muon optimizer \citep{jordan2024muon}, which enables a larger critical batch size and has higher sample efficiency than the widely used AdamW optimizer \citep{loshchilov2019decoupledweightdecayregularization}. Trinity Nano and Trinity Mini were pre-trained on 10 trillion tokens, and Trinity Large was pre-trained on 17 trillion tokens. Additionally, training stability was a key concern in the design of the architecture, due to the limited time and compute availability for the project.

Trinity Large has 400B total parameters, with 13B activated per token, and is trained on a large mixed corpus combining curated web-scale data and synthetic data. In the remainder of this report, we describe the design decisions that govern architecture, training, and post-training, and we evaluate Trinity Large Base as well as Trinity Large Preview across a wide range of standard benchmarks.

\section{Architecture}
\begin{figure}
    \centering
    \includegraphics[width=0.75\linewidth]{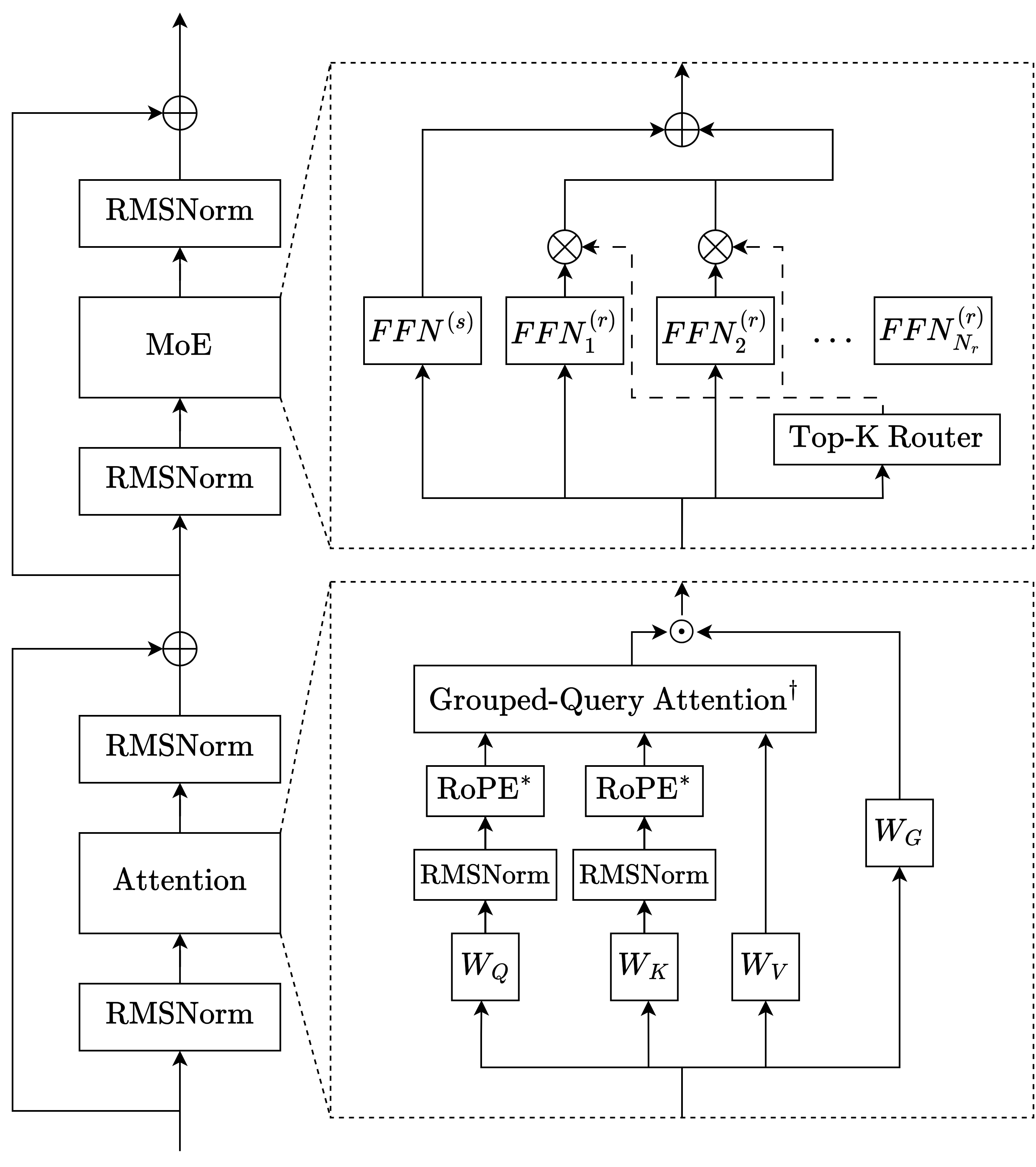}
    \caption{The architecture of the Trinity model family.\newline$^*$ RoPE is only present in local layers.\newline$^\dagger$ The grouped-query attention has a sliding window for the local layers.}
    \label{fig:arch}
\end{figure}
The Trinity models are decoder-only sparse Mixture-of-Experts (MoE) \citep{shazeer2017outrageouslylargeneuralnetworks} transformer models \citep{vaswani2023attentionneed}, with no biases on the linear layers. We discuss the components of the architecture below, including the details of our expert balancing scheme.

\subsection{Tokenizer}

We train a custom 200,000-token BPE \citep{bpeshibata} vocabulary using the \texttt{tokenizers} library \citep{Moi_HuggingFace_s_Tokenizers_2023} on approximately 48GB ($\sim$10B tokens) of training data. This sample was drawn primarily from the pretraining corpus for Trinity Nano and Mini, supplemented with multilingual text from C4 non-English splits and a mix of instruction-following, reasoning-trace, and code data. The larger multilingual corpus used for Trinity Large was not yet finalized at the time of tokenizer training; consequently, non-English languages are less well represented relative to their proportion in the final pretraining data.

\subsubsection{Pretokenization}

Our pretokenizer employs a multi-stage pipeline inspired primarily by the DeepSeek V3 tokenizer \citep{deepseekai2025deepseekv3technicalreport}, with significant extensions for numerical tokenization and multilingual coverage. The pipeline proceeds as follows:

\begin{enumerate}
    \item \textbf{Digit isolation and place-aligned chunking.} Contiguous digit runs are first separated from surrounding text, then split into right-aligned groups of three with a leading group of one to two digits as needed (e.g., \texttt{1234567} $\rightarrow$ \texttt{1|234|567}). This ensures each three-digit token consistently represents a fixed place value (ones-tens-hundreds, thousands, etc.), following the finding of \citet{singh2024tokenizationcountsimpacttokenization} that place-aligned digit tokenization materially improves arithmetic performance in language models. We initially adopted the zero-width lookahead regex from \citet{liu2025superbpespacetravellanguage} (\verb|(?=(\d{3})+(?!\d))|), but discovered during early training runs that this pattern exhibits catastrophic backtracking on documents containing long uninterrupted digit strings, causing tokenization of individual samples to spike to minutes. We addressed this by replacing the single regex with a three-stage pipeline that first caps digit runs at 510 characters, then peels off a leading group of 1--2 digits, and finally chunks the remainder into exact groups of three. The two approaches produce identical splits on all inputs up to 510 digits; the multi-stage version simply avoids the pathological backtracking behavior. As the resulting split boundaries are equivalent, this was applied to the pretokenizer configuration post-hoc without requiring retraining.

    \item \textbf{Script-aware isolation.} We extend DeepSeek V3's CJK isolation to additionally cover Thai, Lao, Khmer, Myanmar, and Korean hangul/jamo --- all scripts that, like CJK, lack whitespace-delimited word boundaries. Isolating these script runs ensures BPE learns merges within each script independently rather than forming cross-script tokens.

    \item \textbf{Word and punctuation splitting.} We adopt the DeepSeek V3 main text regex directly, which handles leading-space word attachment, punctuation-prefixed tokens (covering contractions, code sigils, and markup), and whitespace/newline normalization.

    \item \textbf{Byte-level fallback.} A byte-level encoding layer ensures full coverage of any byte sequence without unknown tokens.
\end{enumerate}

\subsubsection{Vocabulary Size}

We selected 200,000 tokens after an empirical comparison across vocabulary sizes. Because BPE merges are deterministic and greedy, truncating a larger vocabulary by merge rank yields identical tokenization to training at the smaller size, so we trained the full 200k vocabulary and derived smaller variants (including 131k) for ablation. Fertility measurements and small-scale loss curves confirmed consistent gains from the larger vocabulary, with the most pronounced improvements in CJK languages and French --- languages most constrained by smaller vocabularies.

We also evaluated the SuperBPE approach of \citet{liu2025superbpespacetravellanguage}, which trains an initial vocabulary with standard whitespace splitting, truncates it, then resumes training without the whitespace constraint to learn multi-word tokens. While our SuperBPE variant achieved substantially better compression --- particularly on English text ($\sim$29\% fewer tokens) and reasoning traces ($\sim$27\% fewer tokens) --- we were unable to reproduce a corresponding improvement in downstream model performance at our experimental scale. Given this null result, we proceeded with standard BPE at 200k vocabulary.

\subsubsection{Tokenizer Efficiency}

Table~\ref{tab:efficiency} reports bytes-per-token (B/T) and characters-per-token (C/T) across languages and domains, compared to several recent tokenizers. Our tokenizer achieves the strongest compression among standard (non-SuperBPE) tokenizers on English (4.84 B/T on C4-en) and French (3.98 B/T), benefiting from the larger vocabulary. CJK compression is competitive but trails DeepSeek V3 and Qwen 3, which we attribute to the training data timing constraint described above. We expect that training on the complete Trinity Large corpus would close the remaining gap with tokenizers from more CJK-heavy training pipelines.

\begin{table}[ht]
    \centering
    \caption{Comparison of Bytes-per-token (B/T) and Characters-per-token (C/T) across languages and domains.}
    \label{tab:efficiency}
    \vspace{0.5em}
    \begin{tabular}{lccccc}
        \toprule
        Dataset & Trinity (200k) & DeepSeek R1 (128k) & Qwen 3 (152k) & Llama 3 (128k) & GPT-OSS (200k) \\
        \midrule
        C4 (en) B/T & \textbf{4.84} & 4.70 & 4.64 & 4.73 & 4.79 \\
        C4 (zh) C/T & 1.38 & \textbf{1.54} & 1.45 & 1.29 & 1.47 \\
        C4 (ja) C/T & 1.39 & \textbf{1.60} & 1.59 & \textbf{1.60} & 1.54 \\
        C4 (ko) C/T & 1.26 & \textbf{1.47} & \textbf{1.47} & 1.68 & 1.72 \\
        C4 (fr) B/T & \textbf{3.98} & 3.53 & 3.34 & 3.49 & 3.97 \\
        Reasoning B/T & 3.67 & 3.61 & 3.51 & 3.59 & 3.61 \\
        \bottomrule
    \end{tabular}
\end{table}

\subsection{Attention}

Our chosen attention mechanism combine multiple modifications to the standard Multi-Head Attention (MHA) \citep{vaswani2023attentionneed} setup that have been shown to be effective in previous work. We combine grouped-query attention (GQA) \citep{ainslie2023gqatraininggeneralizedmultiquery}, QK-normalization (QK-norm) \citep{henry2020querykeynormalizationtransformers}, gated attention \citep{qiu2025gatedattentionlargelanguage}, and a 3:1 local/global layer pattern, with rotary positional embeddings (RoPE) \citep{su2023roformerenhancedtransformerrotary} and sliding window attention (SWA) \citep{jiang2023mistral7b, gemmateam2024gemma2improvingopen, beltagy2020longformerlongdocumenttransformer, child2019generatinglongsequencessparse} in the local layers, and no positional embeddings (NoPE) (explored by \cite{kazemnejad2023impactpositionalencodinglength}) in the global layers. The local/global arrangement closely follows the results of \cite{yang2025ropenopeagainnew}.

Let $d$ denote the model dimension, $h_q$ the number of query heads, $h_{kv}$ the number of key/value heads, and $d_h = d/h_q$ the per-query-head dimension.
Let $\mathbf{x}_t \in \mathbb{R}^d$ be the input at token position $t$ in a given attention layer.

\vspace{0.25em}
\noindent\textbf{QK-normalization.}
We compute the queries, keys, and values as linear projections of the input and apply RMSNorm to queries and keys (QK-norm) before scaled dot-product attention:
\begin{align}
\mathbf{q}^{0}_{t,i} &= W^{Q}_{i}\mathbf{x}_t, \qquad i\in\{1,\dots,h_q\}, \tag{1}\\
\mathbf{k}^{0}_{t,j} &= W^{K}_{j}\mathbf{x}_t, \qquad j\in\{1,\dots,h_{kv}\}, \tag{2}\\
\mathbf{v}^{0}_{t,j} &= W^{V}_{j}\mathbf{x}_t, \qquad j\in\{1,\dots,h_{kv}\}, \tag{3}\\
\mathbf{q}_{t,i} &= \mathrm{RMSNorm}\!\left(\mathbf{q}^{0}_{t,i}\right), \tag{4}\\
\mathbf{k}_{t,j} &= \mathrm{RMSNorm}\!\left(\mathbf{k}^{0}_{t,j}\right). \tag{5}
\end{align}

where $W^Q_i\in\mathbb{R}^{d_h\times d}$ and $W^K_j,W^V_j\in\mathbb{R}^{d_h\times d}$. We choose to use QK-Norm primarily for training stability reasons, especially due to our use of the Muon optimizer, as prior work has identified growing maximum attention logit values to be more of a concern when using Muon as opposed to AdamW (\cite{liu2025muonscalablellmtraining}; \cite{kimiteam2025kimik2openagentic}). Furthermore, \cite{5team2025glm45agenticreasoningcoding} report that QK-Norm effectively stabilizes the range of maximum attention logit values through a full training run.

\vspace{0.25em}
\noindent\textbf{Local/global attention pattern.}
All models use a 3:1 ratio of local:global attention; they have three local (SWA) attention layers with RoPE, followed by one global layer without positional embeddings (NoPE), with this structure repeating for the full depth of the model. This follows the results of \cite{yang2025ropenopeagainnew}, though we use QK-norm in our final setup. We choose to use this configuration primarily for effective long-context performance as well as for large efficiency gains at longer sequence lengths. The combination of this scheme and gated attention allowed the model to recover performance in a comparatively lower number of steps when training at longer sequence lengths, and also resulted in observed length extrapolation for Trinity Large.

Let $w$ be the local window size. The queries and keys used in the scaled dot-product attention operation are:
\begin{align}
\widehat{\mathbf{q}}_{t,i} &=
\begin{cases}
\mathrm{RoPE}\!\left(\mathbf{q}_{t,i}\right), & \text{local layer},\\
\mathbf{q}_{t,i}, & \text{global (NoPE) layer},
\end{cases} \tag{6}\\
\widehat{\mathbf{k}}_{t,j} &=
\begin{cases}
\mathrm{RoPE}\!\left(\mathbf{k}_{t,j}\right), & \text{local layer},\\
\mathbf{k}_{t,j}, & \text{global (NoPE) layer},
\end{cases} \tag{7}
\end{align}

with valid attention positions:
\begin{align}
\mathcal{S}_t &=
\begin{cases}
\{\,s \mid \max(1,t-w+1)\le s \le t\,\}, & \text{local layer},\\
\{\,s \mid 1 \le s \le t\,\}, & \text{global layer}.
\end{cases} \tag{8}
\end{align}

\vspace{0.25em}
\noindent\textbf{Scaled dot-product attention with GQA.}
During attention, GQA maps each query head $i$ to a key/value head index
\[
j(i) \;=\; \left\lceil i\cdot \frac{h_{kv}}{h_q} \right\rceil,
\]
such that multiple query heads share the same key/value head. This has been shown to roughly match the performance of MHA while drastically reducing KV-cache size. 

For each query head $i$, we attend over keys/values from the shared KV head $j(i)$:
\begin{align}
\alpha_{t,i,s} &=
\mathrm{Softmax}_{s\in \mathcal{S}_t}
\left(
\frac{\widehat{\mathbf{q}}_{t,i}^{\top}\widehat{\mathbf{k}}_{s,j(i)}}{\sqrt{d_h}}
\right), \tag{9}\\
\mathbf{o}^{\mathrm{sdpa}}_{t,i} &= \sum_{s\in \mathcal{S}_t} \alpha_{t,i,s}\,\mathbf{v}_{s,j(i)}. \tag{10}
\end{align}

\vspace{0.25em}
\noindent\textbf{Gated attention.}
Following a configuration from \cite{qiu2025gatedattentionlargelanguage}, we apply an elementwise gating to the scaled dot-product attention output before the output linear projection:
\begin{align}
\mathbf{g}_t &= \sigma\!\left(W^{G}\mathbf{x}_t\right), \tag{11}\\
\mathbf{g}_{t,i} &= \mathrm{split}_{h_q}\!\left(\mathbf{g}_t\right)_{i}, \tag{12}\\
\widetilde{\mathbf{o}}_{t,i} &= \mathbf{o}^{\mathrm{sdpa}}_{t,i}\odot \mathbf{g}_{t,i}, \tag{13}\\
\mathbf{u}_t &= W^{O}\,[\widetilde{\mathbf{o}}_{t,1};\widetilde{\mathbf{o}}_{t,2};\dots;\widetilde{\mathbf{o}}_{t,h_q}]. \tag{14}
\end{align}

where $\sigma(\cdot)$ is the sigmoid, $\mathrm{split}_{h_q}(\cdot)$ partitions $\mathbf{g}_t\in\mathbb{R}^{d}$ into $h_q$ contiguous vectors in $\mathbb{R}^{d_h}$, and $W^{G}\in\mathbb{R}^{d\times d}$, $W^{O}\in\mathbb{R}^{d\times d}$ are the gate and output projections, respectively. In \cite{qiu2025gatedattentionlargelanguage}, gated attention was identified to reduce attention sinks, reduce overly large activations, improve performance on evaluations, and improve long-sequence generalization. Furthermore, gating appears to stabilize training and reduce the occurrence of loss spikes during the training process, which is also a key motivation for our adoption.

\subsection{Mixture-of-Experts}

Our Mixture-of-Experts layers follow the DeepSeekMoE \citep{dai2024deepseekmoeultimateexpertspecialization} design, with fine-grained routed experts and an always-active shared expert. We use the SwiGLU \citep{shazeer2020gluvariantsimprovetransformer} activation function as the nonlinearity. For Trinity Large, in order to facilitate better throughput, we opted to use coarser-grained experts. We also replace the first $k$ MoE layers with dense layers to stabilize early representations. We use sigmoid routing following \cite{wang2024auxiliarylossfreeloadbalancingstrategy}, normalizing the router scores before using them for gating, and we also apply the gating scores to the output of each expert.
    
\noindent\textbf{MoE formulation.}
Let $\mathbf{u}_t\in\mathbb{R}^{d}$ denote the MoE input at token $t$.
The output of the MoE module $\mathbf{h}'_t$ is computed as:
\begin{align}
\mathbf{h}'_t
&=
\mathbf{u}_t
+
\sum_{i=1}^{N_s}\mathrm{FFN}^{(s)}_{i}\!\left(\mathbf{u}_t\right)
+
\sum_{i=1}^{N_r} g_{i,t}\,\mathrm{FFN}^{(r)}_{i}\!\left(\mathbf{u}_t\right).
\tag{15}
\end{align}

where $\mathrm{FFN}^{(s)}_i(\cdot)$ denotes the $i$-th shared expert and $\mathrm{FFN}^{(r)}_i(\cdot)$ denotes the $i$-th routed expert.

\noindent\textbf{Routing.}
We use normalized sigmoid routing scores instead of softmax-based routing, which allows for more stable router logits. We also follow \cite{wang2024auxiliarylossfreeloadbalancingstrategy} in using the sum of the routing scores and expert bias to determine the Top-$K$ experts, but using the routing scores without expert bias as gating scores for the outputs of the routed experts, since the expert bias is updated in a decoupled way.
Let $\mathbf{e}_i\in\mathbb{R}^{d}$ be the router vector for routed expert $i$:
\begin{align}
s_{i,t}
&=
\sigma\!\left(\mathbf{u}_t^{\top}\mathbf{e}_i\right),
\qquad i\in\{1,\dots,N_r\}.
\tag{16}
\end{align}

The Top-$K$ experts are selected as follows:
\begin{align}
g'_{i,t}
&=
\begin{cases}
s_{i,t},
&
s_{i,t} + b_i \in \mathrm{TopK}\!\left(\{s_{j,t}+b_j\}_{j=1}^{N_r},\,K_r\right),
\\[0.2em]
0,
&
\text{otherwise},
\end{cases}
\tag{17}\\
g_{i,t}
&=
\frac{g'_{i,t}}{\sum_{j=1}^{N_r} g'_{j,t}}.
\tag{18}
\end{align}

where $b_i$ is the expert bias, and $g_{i,t}$ are the gating scores derived from the router scores $s_{i,t}$.

\noindent\textbf{Auxiliary-loss-free load balancing.}
For Trinity Mini and Trinity Nano, we used the standard formulation of auxiliary-loss-free load balancing \citep{wang2024auxiliarylossfreeloadbalancingstrategy}, with an additional re-centering of the expert bias updates.
We maintain a bias vector $\mathbf{b}=[b_1,\dots,b_{N_r}]$ that is updated in a decoupled fashion.
Let $n_i$ be the number of tokens routed to expert $i$ in the current step, and $\bar{n}$ the mean load:
\begin{align}
\bar{n}
&=
\frac{1}{N_r}\sum_{i=1}^{N_r} n_i,
\tag{19}\\
\Delta b_i
&=
\gamma \cdot \mathrm{sign}\!\left(\bar{n}-n_i\right),
\tag{20}\\
b_i
&=
b_i + \Delta b_i,
\tag{21}\\
b_i
&=
b_i - \frac{1}{N_r}\sum_{j=1}^{N_r} b_j.
\tag{22}
\end{align}

where $\gamma$ is the bias update speed.

\noindent\textbf{Soft-clamped Momentum Expert Bias Updates (SMEBU) load balancing.}
For Trinity Large, we replace the per-step sign-based expert bias update with a $\tanh$ soft-clamped magnitude-aware update.
First, we compute the normalized per-expert violation, in order to make the update scale independent of sequence-length and batch size, and apply $\tanh$ soft-clamping:
\begin{align}
v_i
&=
\frac{\bar{n}-n_i}{\bar{n}},
\tag{23}\\
\tilde{v}_i
&=
\tanh\!\left(\kappa\,v_i\right),
\tag{24}
\end{align}

where $\kappa$ is a tunable scale.
We center the update around zero, and additionally use a momentum buffer to smooth expert bias updates over time:
\begin{align}
\Delta b_i
&=
\lambda\,\tilde{v}_i,
\tag{25}\\
\Delta b_i
&=
\Delta b_i - \frac{1}{N_r}\sum_{j=1}^{N_r}\Delta b_j,
\tag{26}\\
m_i
&=
\beta\,m_i + (1-\beta)\,\Delta b_i,
\tag{27}\\
b_i
&=
b_i + m_i,
\tag{28}
\end{align}

where $\lambda$ is the load-balance learning rate, $\beta$ is the momentum factor, and $\mathbf{m}=[m_1,\dots,m_{N_r}]$ is the maintained momentum buffer.

When training Trinity Large, our initial runs faced router instability issues, and as one method of stabilizing the run, we examined the load-balancing method. Under the assumption that the ideal expert bias value is a fixed value, we note that the standard aux-free load balancing cannot precisely converge on that value, as each local update under the $\mathrm{sign}(\cdot)$ operator is always $\pm \lambda$. We hypothesize that this is a cause of instability for MoE training with the standard aux-loss-free objective. As the total number of experts increases, the per-layer norm of the bias step also increases. Viewing this as an optimization process, as this nears its local minima, the update step is biased towards oscillation, and can never fully settle.

We alleviate this using $\tanh$ as a continuous relaxation of the discrete update that can converge to 0. We replace the $\mathrm{sign}(\cdot)$ operator with $\tanh$ as a smooth approximation, and add a tunable scale $\kappa$ to control saturation speed. We believe that bounded updates are important to maintaining network stability; preliminary tests using a linear, unclamped update quickly reduced MaxVio early on in training but resulted in instability later in training. Additionally, consistent with the hypothesis that behavior near convergence is one of the causes of instability in MoE aux-loss-free load balancing, and that noise near a local minima should be roughly decorrelated over time, we introduce momentum as a form of noise dampening. This is analogous to momentum SGD reducing variance in noisy gradient updates.
For our Trinity Large run, we use SMEBU with $\lambda = 5\times10^{-4}$, $\beta = 0.5$, and $\kappa = 2$.

\newpage
\noindent\textbf{Sequence-wise auxiliary loss.}
In addition to aux-loss-free balancing, we use a complementary sequence-wise load balance loss to promote balance within a sequence, following the formulation from \cite{deepseekai2025deepseekv3technicalreport}:
\begin{align}
\mathcal{L}_{\mathrm{Bal}}
&=
\alpha \sum_{i=1}^{N_r} f_i P_i,
\tag{29}\\
f_i
&=
\frac{N_r}{K_r T}\sum_{t=1}^{T}
\mathbbm{1}\!\left(
s_{i,t}+b_i \in \mathrm{TopK}\!\left(\{s_{j,t}+b_j\}_{j=1}^{N_r},\,K_r\right)
\right),
\tag{30}\\
\tilde{s}_{i,t}
&=
\frac{s_{i,t}}{\sum_{j=1}^{N_r} s_{j,t}},
\tag{31}\\
P_i
&=
\frac{1}{T}\sum_{t=1}^{T}\tilde{s}_{i,t},
\tag{32}
\end{align}

where $\alpha$ is a small coefficient, $T$ is the sequence length, and $\mathbbm{1}(\cdot)$ is the indicator function.

\subsection{Normalization}
For layer normalization, we use a simplified depth-scaled sandwich norm \citep{yin2025panguultrapushinglimits, ding2021cogviewmasteringtexttoimagegeneration, kim2025perilnrevisitingnormalizationlayer}. Both the input and output of the module are normalized:
\begin{align}
\mathbf{y}_\ell
&=
\mathbf{x}_\ell
+
\mathrm{RMSNorm}^{(2)}_\ell\!\Big(
\mathcal{M}_\ell\big(
\mathrm{RMSNorm}^{(1)}_\ell(\mathbf{x}_\ell)
\big)
\Big),
\tag{33}
\end{align}

where $\mathbf{x}_\ell$ and $\mathbf{y}_\ell$ are the input/output of layer $\ell$, and $\mathcal{M}_\ell(\cdot)$ is the sublayer module (attention, FFN, or MoE).

We depth-scale the RMSNorm gain parameters for the second RMSNorm in each layer.
Let $L$ be the total number of layers and $\gamma(\cdot)$ denote the RMSNorm gain; we initialize:
\begin{align}
\gamma\!\left(\mathrm{RMSNorm}^{(1)}_\ell\right)
&=
1,
\tag{34}\\
\gamma\!\left(\mathrm{RMSNorm}^{(2)}_\ell\right)
&=
\frac{1}{\sqrt{L}}.
\tag{35}
\end{align}

We also apply RMSNorm before the language modeling head:
\begin{align}
\mathbf{z}
&=
\mathrm{RMSNorm}_{\mathrm{LM}}(\mathbf{h}_L).
\tag{36}
\end{align}

\subsection{Initialization}
All trainable parameters are initialized from a zero-mean truncated normal distribution, with width-scaled standard deviation:
\begin{align}
\theta &\sim \mathrm{TruncNormal}\!\left(0,\;\sigma^2;\;[-3\sigma,\;3\sigma]\right),
\tag{37}\\
\sigma &= \frac{0.5}{\sqrt{d}},
\tag{38}
\end{align}

where $d$ is the model dimension. This roughly follows the findings in \cite{takase2025spikemorestabilizingpretraining}, which recommends setting $\sigma = \sqrt{\frac{2}{5d}}$. We note that $\frac{0.5}{\sqrt{d}}$ aligns with the initialization scale for DeepSeek-V3 \citep{deepseekai2025deepseekv3technicalreport}, as $\frac{0.5}{\sqrt{7168}}=0.006$.

During the forward pass, we scale the embedding layer's activations by $\sqrt{d}$, also following \cite{takase2025spikemorestabilizingpretraining}:
\begin{align}
\mathbf{e}_t &= \sqrt{d}\;E(\mathrm{tok}_t).
\tag{39}
\end{align}
We also note that both the Grok-1 and the Grok-2 model checkpoints on HuggingFace \citep{xai_grok1_hf, xai_grok2_hf} have \texttt{embedding\_multiplier\_scale} set to $\sqrt{d}$. Additionally, the first two generations of the Gemma models \citep{gemmateam2024gemmaopenmodelsbased, gemmateam2024gemma2improvingopen} refer to this multiplier as a normalizer in their modeling code for HuggingFace's \texttt{transformers} library \citep{huggingface_transformers_github}.

\section{Pre-training}

\subsection{Data}

All pretraining data were curated by DatologyAI. Trinity Nano and Trinity Mini were each trained on the same 10 trillion token mix, and Trinity Large was trained on 17 trillion tokens of a distinct 20 trillion token mix. Both mixes were organized into three phases. The 10 trillion mix was distributed across 7T tokens in phase 1, 1.8T tokens in phase 2, and 1.2T tokens in phase 3. The 20 trillion mix was distributed across 13T tokens in phase 1, 4T tokens in phase 2, and 3T tokens in phase 3. The 17 trillion tokens that Trinity Large was trained on were sampled proportionally from the 20T mix (i.e. a 13:4:3 ratio across the three phases). Each mix was designed to target both general English capabilities as well as programming, STEM, and reasoning skills. The 10 trillion mix reuses the AFM-4.5B dataset and adds substantially more math and code. The 20T Trinity Large mix uses state-of-the art programming, STEM, reasoning, and multilingual data curation, targeting Arabic, Mandarin, Japanese, Spanish, German, French, Italian, Portuguese, Indonesian, Russian, Vietnamese, Hindi, Korean, and Bengali.

A key component of this curation is large-scale synthetic data generation. DatologyAI generated over 8 trillion tokens of synthetic web, code, and STEM data. For the web, approximately 6.5 trillion tokens of synthetic data were generated using rephrasing approaches that build on top of BeyondWeb \citep{maini2025beyondweb}. Briefly, high-quality seed documents were selected from web corpora and then diverse generation strategies were used to rephrase these documents, including format transformation (e.g. converting web content into question–answer pairs), style modification (e.g. enhancing pedagogical tone), and content restructuring to improve information density and accessibility. Similar approaches were used to generate approximately 1 trillion tokens of synthetic multilingual data.

Approximately 800 billion tokens of high quality synthetic code data were also generated. In a similar fashion to web, relevant high-quality code files were selected for rephrasing and enhancement using a breadth of approaches that maximize diversity and relevance to task and style.

Efficiently and reliably generating over 8 trillion tokens of synthetic data requires robust, scalable infrastructure. The DatologyAI curation stack utilizes Ray and vLLM on Kubernetes to support heterogeneous clusters, various GPU types, and seamless scaling.

The data mixes for the Trinity models all performed midtraining by shifting the data mix in 3 phases towards higher-quality and domain specific data as discussed in \citep{blakeney2024does, hu2024minicpm, olmo20252olmo2furious, allal2025smollm2}. This included both shifting the data mix away from general web to more code, math, and science over the course of the phases as well as shifting to higher quality and more relevant data within these categories, e.g. within web, code, and STEM.

To our knowledge, this represents one of the largest publicly documented efforts in synthetic data generation for pretraining, demonstrating that carefully curated synthetic data can be produced at the multi-trillion token scale required for frontier model development. The effectiveness of the curation approach as a whole is reflected in the downstream evaluations presented in Section \ref{sec:evals}, where the Trinity models demonstrate strong performance across the targeted capability domains.

\subsection{Data Preparation}

Our pre-training data has three phases. Each phase consists of the same mixture, but the distribution of math and code data is boosted in later phases, in order to increase the final performance of the model. We tokenize documents on the fly and use sequence packing to construct token sequences with the remaining tokens from the document stream accumulated in a buffer for the next sample, with multiple dataloader workers used to interleave documents across batches. For Trinity Nano and Trinity Mini, we do not use intra-doc attention masking, but Trinity Large was pre-trained using intra-doc attention masking. During the pre-training of Trinity Large, we were concerned that our method of on-the-fly tokenization could cause longer documents to remain in consecutive batches for too long, due to insufficient interleaving. To reduce this inter-batch correlation, we introduce a document buffer mid-way through the Trinity Large training run, right before beginning phase 3.

One standard approach to sequential packing for on-the-fly tokenization, as implemented in frameworks like TorchTitan \citep{liang2025torchtitan}, continues documents that span multiple times the sequence length into subsequent batch items. While straightforward, this can become problematic when document lengths follow a lognormal distribution, as long documents introduce unbalanced domain biases at the minibatch level. This creates overhead in the learning objective as noise from step-to-step distribution imbalances must be corrected at every training step. We note that this is a consequence of sequential packing, and cannot simply be avoided by shuffling input documents.

We hypothesize that imbalanced minibatches have particular impact in at least two scenarios: (i) in limited-batchsize settings, and (ii) in regimes where the model becomes more data efficient per step. In both settings, informally, if we treat each unbalanced minibatch as an ideally-representative sample from its own target data generator, we can see that we effectively optimize a different target distribution per step. This not only causes additional overhead, as the network must ``unlearn'' the small domain imbalances every step, but also introduces a source of long-tail noise due to the lognormal distribution of sequence lengths. Under asymmetric losses like cross-entropy, small deviations can result in larger loss fluctuations. Additionally, as models grow larger, their step-to-step data efficiency tends to increase \citep{kaplan2020scalinglawsneurallanguage}, which we hypothesize makes them more vulnerable to the step-to-step fluctuations in the data distribution. We hypothesize that this is one potential contributor to training instability as models grow larger.

To reduce this intra-batch correlation, we introduce a new method called the Random Sequential Document Buffer (RSDB). The Random Sequential Document Buffer operates as follows: after tokenizing a document, we load the entire token sequence as an entry in the RSDB, and initialize a read head at the $0$th index. We continue tokenizing and inserting until the RSDB is full. When populating a single sequence buffer within a minibatch in the dataloader, until the sequence buffer is full, we randomly sample an index of the document in the RSDB, and read tokens from the document at the position of the read head into the sequence buffer. Document read head positions are updated the current position in the document after every step. If the sequence buffer is full, we return it, if not, we randomly select another document index and continue to read tokens into the sequence buffer. This process is continued until the sequence buffer is full. The buffer is refilled in order to keep a sufficient number of active documents.

For runtime efficiency, we set the internal buffer size to twice the user-specified buffer value, and refill the buffer to this larger value as soon as the buffer reaches the user-specified value. We also purge old documents and load new documents in bulk during this refill step. We found this optimization significantly improved dataloader performance. For phase 3 of training for Trinity Large, we used an RSDB with a buffer size of $8192$ per GPU (user-specified $4096$), with 4 workers per GPU. We split the RSDB evenly across each worker to keep the total size of the RSDB independent of worker counts.

The process addresses the previously mentioned issues in several ways. First, it reduces fragmentation by not pre-chunking documents, which could result in more fragmentation with an online algorithm, instead dynamically packing them at runtime. Second, it retains truly random selection across documents, better respecting IID sampling properties with respect to the data distribution. Third, it is memory read/write efficient, allowing for an efficient implementation that does not bottleneck the training process.

As an efficient way to quantify batch imbalance during training, and the efficacy of our method, we introduce a metric called Batch Heterogeneity, (BatchHet), defined as the difference between the maximum microbatch loss and the average loss for each training step:

\begin{align}
\mathrm{BatchHet}_t &= \max_{i} \mathcal{L}_i^{(t)} - \frac{1}{M} \sum_{i=1}^{M} \mathcal{L}_i^{(t)}
\tag{40}
\end{align}

where $\mathcal{L}_i^{(t)}$ is the loss of microbatch $i$ at training step $t$, and $\bar{\mathcal{L}}^{(t)} = \frac{1}{M} \sum_{i=1}^{M} \mathcal{L}_i^{(t)}$ is the mean loss across all $M$ microbatches at that step.

In small scale internal experiments, we observe that this metric correlates strongly with gradient norm instability. In a small internal experimental setup, enabling RSDB reduced BatchHet by $46\%$ and improved loss over a sequential packing baseline. Additionally, the gradient norm was markedly stable, with a kurtosis of $14.6$ for the grad norm of the RSDB-enabled network, as opposed to the kurtosis of $187$ of the baseline's grad norm. Matching the step-to-step loss variance of the RSDB-enabled network required roughly a $2\times$ increase in the batchsize of the baseline network, due to increased data inefficiency. However, even with the loss-variance-matched batchsize, the baseline network had significantly higher step-to-step outliers in the average loss value when compared to the RSDB-enabled networks. A batchsize increase of $7\times$ was required in order for the BatchHet of the baseline to match the BatchHet of the RSDB-enabled network. We note that these improvements happen without dropping any tokens.

During development of the RSDB, we noted significant enough performance gains from it that we decided to integrate it during phase 3 of the Trinity Large training run instead of waiting for a later training run. While the data distributions between phase 2 and phase 3 make direct comparison difficult, the overall effect was notable: BatchHet reduced by a factor of 4.23x, and step-to-step variance reduced by a factor of 2.4x (see Figure \ref{fig:loss}), a significant improvement when compared to the default packing strategy. We note that training runs without the RSDB exhibit much higher values in the higher-order moments of the running loss distribution, which we believe to correlate with network instability during training.

\subsection{Infrastructure}

We trained Trinity on GPU clusters managed by Prime Intellect. We use a modified version of TorchTitan \citep{liang2025torchtitan} as our training framework. For efficient fused cross-entropy and z-loss, we use kernels from the Liger Kernels \citep{hsu2025ligerkernel} project. For larger memory savings during context extension and supervised fine-tuning, we use Cut Cross-Entropy \citep{wijmans2025cut}.

Trinity Nano and Trinity Mini were trained on clusters of 512 H200 GPUs, and Trinity Large was trained on a cluster of 2048 B300 GPUs. We train Trinity Nano and Trinity Mini with a Hybrid Sharded Data Parallel (HSDP) configuration, where there are multiple replicas of the model and Fully-Sharded Data Parallel (FSDP) \citep{zhao2023pytorchfsdpexperiencesscaling} is used within the replica group. For Trinity Large, we additionally employ Expert Parallelism (EP) within a GPU node for efficient Mixture-of-Experts Layers. 

For all models, we set the FSDP group size to 128. For Trinity Large, we set the EP group size to 8. For the context extension of Trinity Large, we use a context parallelism degree of 4.

We use a modification of the Muon implementation in the Dion repository \citep{ahn2025dion} for efficient distributed Muon. We orthogonalize the gradient for the experts in a batched fashion, without flattening the gradient tensor prior to orthogonalization. This also simplifies the expert-parallel implementation, since gradients do not need to be gathered across the expert-parallel group.

During the training of Trinity Large, when testing out versions of the PyTorch 2.10 nightly, we came across an issue where early versions did not fully support B300s, but later versions had a performance regression for both B200s and B300s. We conducted a binary search across nightly versions to find a package that worked well for the training setup. Later in the run, we upgraded to a recent PyTorch 2.11 nightly, giving us roughly a 5\% speedup.

We designed the training stack to recover quickly from hardware failures. Since Trinity Large was trained on brand new B300 systems and is among the first large scale runs on this hardware, we expected more frequent failure rates during early deployment. In practice, we observed intermittent GPU faults, including recurrent XID errors, which were gradually mitigated through updated firmware bundles. On the systems side, we use failover nodes and prioritize fast restarts through dataloader optimizations and high performance NFS, significantly reducing recovery time after training restarts. We also run continuous heartbeat monitoring that triggers alerts if no training step completes within a three minute window, allowing us to quickly detect stalled jobs and intervene.

\subsection{Hyperparameters}

\subsubsection{Model Configurations}

\begin{table}[t]
\centering
\setlength{\tabcolsep}{6pt}
\caption{Model configurations for the Trinity model variants.}
\label{tab:trinity_specs}
\begin{tabular}{lccc}
\toprule
\textbf{} & Trinity Nano & Trinity Mini & Trinity Large \\
\midrule
Transformer layers & 56 & 32 & 60 \\
Initial dense layers & 2 & 2 & 6 \\
Model dim ($d_{\text{model}}$) & 1024 & 2048 & 3072 \\
FFN intermediate dim & 3072 & 6144 & 12288 \\
Attention heads ($h_q$) & 8 & 32 & 48 \\
Per-head dim ($d_h$) & 128 & 128 & 128 \\
KV heads ($h_{kv}$) & 2 & 4 & 8 \\
Local window size & 2048 & 2048 & 4096 \\
Pre-training seq len & 4096 & 4096 & 8192 \\
MoE shared experts & 1 & 1 & 1 \\
MoE routed experts & 128 & 128 & 256 \\
Activated experts / token & 8 & 8 & 4 \\
Route scale & 2.826 & 2.826 & 2.448 \\
Expert size & 256 & 1024 & 3072 \\
Initialization (trunc normal $\sigma$) & 0.016 & 0.011 & 0.009 \\
\bottomrule
\end{tabular}
\end{table}

\textbf{Trinity Nano.}
We set the number of Transformer layers to $56$ with $2$ initial dense layers, and the model dimension to $1024$ with FFN intermediate dimension $3072$.
For attention, we use $8$ heads with per-head dimension $128$ ($h_q{=}8$, $d_h{=}128$) and $2$ KV heads ($h_{kv}{=}2$).
We use local/global attention with sliding-window size for the local layers set to $2048$ and pre-training sequence length $4096$.
For each MoE layer we use $1$ shared expert and $128$ routed experts, activating $8$ experts per token, with route scale set to $2.826$. The expert size was set to $256$.
We initialize all trainable parameters from a truncated normal distribution with standard deviation $0.016$.

With Trinity Nano, we aimed to explore the effectiveness of high depth at producing a powerful small language model, following the improved performance of Falcon-H1-1.5B-Deep over Falcon-H1-1.5B \citep{zuo2025falconh1familyhybridheadlanguage}.

\textbf{Trinity Mini.}
We set the number of Transformer layers to $32$ with $2$ initial dense layers, and the model dimension to $2048$ with FFN intermediate dimension $6144$.
For attention, we use $32$ heads with per-head dimension $128$ ($h_q{=}32$, $d_h{=}128$) and $4$ KV heads ($h_{kv}{=}4$).
We use local/global attention with sliding-window size for the local layers set to $2048$ and pre-training sequence length $4096$.
For each MoE layer we use $1$ shared expert and $128$ routed experts, activating $8$ routed experts per token, with route scale set to $2.826$. The expert size was set to $1024$.
We initialize all trainable parameters from a truncated normal distribution with standard deviation $0.011$.

\textbf{Trinity Large.}
We set the number of Transformer layers to $60$ with $6$ initial dense layers, and the model dimension to $3072$ with FFN intermediate dimension $12288$.
For attention, we use $48$ heads with per-head dimension $128$ ($h_q{=}48$, $d_h{=}128$) and $8$ KV heads ($h_{kv}{=}8$).
We use local/global attention with sliding-window size for the local layers set to $4096$ and pre-training sequence length $8192$.
For each MoE layer we use $1$ shared expert and $256$ routed experts, activating $4$ experts per token, with route scale set to $2.448$. The expert size was set to $3072$. For Trinity Large, we opted to activate 4 routed experts and make each expert larger, reducing the granularity, due to throughput requirements. We also greatly increase the sparsity to maximize model capacity while keeping a low active parameter count.
We initialize all trainable parameters from a truncated normal distribution with standard deviation $0.009$.

\subsubsection{Training Hyperparameters}
When training all three models, we use the Muon optimizer \citep{jordan2024muon} for hidden layers and the AdamW optimizer \citep{loshchilov2019decoupledweightdecayregularization} for the embedding and output layers. Unlike \cite{liu2025muonscalablellmtraining}, we choose not to rescale the RMS of the Muon updates to match the RMS of the AdamW updates. Instead, we use the learning rate adjustment rule:
\begin{align}
\mathrm{lr}_{\mathrm{adjusted}}
&=
\mathrm{lr}\,\sqrt{\max\!\left(1,\frac{\mathrm{fan}_{\mathrm{out}}}{\mathrm{fan}_{\mathrm{in}}}\right)}.
\tag{41}
\end{align}

following \cite{jordan2024muon_repo}. Empirically, we observe that this adjustment enables optimal learning rate transfer when scaling model width, for Muon. 

\textbf{Trinity Nano.}
We use a linear warmup of $2000$ steps.
The peak learning rates are $1.0\times 10^{-3}$ for Muon and $3.0\times 10^{-4}$ for AdamW.
We train with global batch size $4096$ at sequence length $4096$, and increase the global batch size to $8192$ when scaling from a $256$-GPU cluster to a $512$-GPU cluster to improve hardware utilization.
After warmup, we apply a linear decay to zero during the decay stage.
For context extension, we use a linear re-warmup to $\tfrac{1}{10}$ of the peak learning rate followed by a linear decay back to zero.
We train to $256\text{k}$ context for inference at $128\text{k}$.

\textbf{Trinity Mini.}
We use a linear warmup of $2000$ steps.
The peak learning rates are $1.0\times 10^{-3}$ for Muon and $2.0\times 10^{-4}$ for AdamW.
We train with global batch size $4096$ at sequence length $4096$, and increase the global batch size to $8192$ when scaling from a $64$-GPU cluster to a $512$-GPU cluster to improve hardware utilization.
After warmup, we apply a linear decay to zero during the decay stage.
For context extension, we use a linear re-warmup to $\tfrac{1}{10}$ of the peak learning rate followed by a linear decay back to zero.
We train to $128\text{k}$ context for inference at $128\text{k}$.

\textbf{Trinity Large.}
We use a linear warmup of $2000$ steps.
The peak learning rates are $8.0\times 10^{-4}$ for Muon and $2.0\times 10^{-4}$ for AdamW.
We train with global batch size $12288$ at sequence length $8192$, and increase the global batch size to $16384$ after crossing $4.9$T tokens to improve throughput, roughly following \cite{minimax2025minimax01scalingfoundationmodels}.
During the decay phase, we use a cosine decay to $\tfrac{1}{10}$ of the peak learning rate, i.e., to $8.0\times 10^{-5}$ (Muon) and $2.0\times 10^{-5}$ (AdamW), which allows us to proceed into context extension without re-warming.
For context extension, we continue cosine decay from $\tfrac{1}{10}$ to $\tfrac{1}{20}$ of the peak learning rate, i.e., from $8.0\times 10^{-5}$ to $4.0\times 10^{-5}$ (Muon) and from $2.0\times 10^{-5}$ to $1.0\times 10^{-5}$ (AdamW).
We train to $256\text{k}$ context for inference at $512\text{k}$.

\subsection{Context Extension}

The Trinity architecture utilizes interleaved local/global attention, with the local attention layers pre-trained at a sliding window size set to half of the context window of global layers. In initial context extension experiments, we compared adjusting the local layers (with different values for the RoPE base frequency $\theta$) along with the global layers against adjusting only the global layers. Adjusting only the global layers resulted in a much quicker loss recovery, aligning with the findings in \cite{yang2025ropenopeagainnew}, and further suggesting that the local and global layers learn complementary roles. Additionally, not extending the window size of the local layers allows for more efficient inference, so we choose to follow this approach to extend the context window of the models. As a result, the context extension process was extremely smooth.

Due to our time and compute constraints, we used a rapid iterative approach for testing context extension effectiveness, using the Multi-Key Needle-in-a-haystack (MK-NIAH) task from RULER \citep{hsieh2024rulerwhatsrealcontext} at the target context length as a proxy metric. This allowed us to rapidly evaluate checkpoints and make decisions about what to adjust for the next iteration. Consistent with \cite{gao2025trainlongcontextlanguagemodels} and \cite{nvidia2025nvidianemotronnano2}, we saw that training at a sequence length longer than the target context window enables better performance at the target context window. When extending the context window of Trinity Nano, training at a sequence length of 128K (the target context window) yielded an MK-NIAH (@128K) score of $0.38$, whereas training at a sequence length of 256K yielded an MK-NIAH (@128K) score of $0.548$. Iterating on and improving our selection of dataset and hyperparameters yielded a final MK-NIAH (@128K) score of $0.864$ for Trinity Nano. Due to resource constraints, we trained Trinity Mini at 128K for a target of 128K, but found that it performed significantly stronger compared to training Trinity Nano at 128K alone. Our final iteration on Trinity Mini, trained at a sequence length of 128K, achieved a final MK-NIAH (@128K) score of $0.888$.

This effect of model size affecting ease of context extension continued for Trinity Large. We trained Trinity Large at a sequence length of 256K, targeting a final context window size of 256K, and the model achieved an MK-NIAH (@256K) score of $0.994$. Furthermore, upon evaluating it at 512K, we even found that it achieved an MK-NIAH (@512K) score of $0.976$, despite not being trained at that sequence length. Additionally, even evaluating it at 1M, it achieved an MK-NIAH (@1M) score of $0.42$, suggesting that it would be possible to push later versions of the model to have strong performance at a context window of 1M. Additionally, we did not find any improvements from progressively increasing the context window size as opposed to training directly at the longest sequence length, beginning from the final pre-trained checkpoint.

Our long-context extension dataset comprises approximately 117 billion tokens across 35.6 million documents. Following established best practices, we blend samples from our original pretraining distribution with targeted long-context sources. For the pretraining component, we apply a length-biased sampling strategy across all three phases of the pre-training corpus: sampling probabilities scale with document length, from a 1\% floor up to 90\% for documents reaching 128K tokens. This approach yields a subset that remains broadly representative of the source distribution while substantially enriching the proportion of longer sequences. Inspired by Allen AI's Dolma 3 Longmino mix for OLMo 3 \citep{olmo2025olmo3}, we incorporate substantial OCR-derived PDF data, including their olmOCR \citep{poznanski2025olmocrunlockingtrillionstokens} dataset and HuggingFace's FinePDF-edu \citep{kydlíček2025_finepdfs_liberating_3t_of_the_finest_tokens_from_pdfs}, which provide naturally long, high-quality documents. We also regenerate the ProLong datasets \citep{gao2025trainlongcontextlanguagemodels} at full sequence length rather than the published 512K/64K truncations; whole-repository code concatenations from this source proved particularly valuable for both extended-context generalization and cross-file code understanding. The mix is rounded out with instruction-style data (FLAN \citep{chung2022scalinginstructionfinetunedlanguagemodels}, math, and code) drawn from our original AFM context-extension recipe to preserve few-shot and instruction-following capabilities, alongside AutoMathText \citep{zhang2025autonomousdataselectionzeroshot} and curated arXiv and textbook sources from ProLong.

\section{Post-Training}

We conducted the post-training for Trinity Large on the same cluster as the pre-training. Because we allocated most of our cluster time to pretraining, we were constrained to a relatively light post-training phase. As a result, the model we report here, Trinity-Large-Preview, is best viewed as a preliminary release rather than a fully post-trained model. In the next iteration, we plan to build upon this foundation with further, more extensive post-training. 

\textbf{Dataset construction.} We build our instruction-tuning mix from a blend of public instruction data and custom data. The custom portion combines human-written prompts with synthetic instructions generated by stronger teacher models. These are then filtered for quality, formatting, and length. We also lean heavily into agentic coding supervision: a large share of our coding subset comes from trajectories collected through agentic harnesses. In particular, we use OpenCode \citep{opencodeai2026} as a harness to run multi-step coding tasks and record full interaction traces (edits, tool calls, and test outcomes), so the model learns the edit-run-test loop rather than only a final completion.

\textbf{Supervised fine-tuning.} For the supervised fine-tuning stage, we also use our modified version TorchTitan as the training framework. This allows us to use a similar parallelism setup to our context extension phase, though without requiring context parallelism. We use the Cut Cross-Entropy kernel to reduce memory requirements. To increase training efficiency, we pre-tokenize samples offline and during training, we use greedy packing to process samples in parallel. We train at a sequence length of 64K for the supervised fine-tuning stage. When tokenizing samples, we truncate any samples longer than the target sequence length, and during training, we pad sequences to the full sequence length after packing. We run supervised fine-tuning using a linear warm-up for the learning rate, up to $\frac{1}{10}$ of the peak pre-training learning rate. This is followed by linearly decaying the learning rate to zero.

\textbf{Reinforcement learning.} After supervised fine-tuning, we run a short RL stage using prime-rl \citep{primeintellectteam2025intellect3technicalreport} as our training framework on the same 2048 B300 cluster. We use prime-rl's asynchronous setup, where vLLM-backed workers generate rollouts while a separate distributed trainer applies updates with FSDP2. The environments we use follow the \texttt{verifiers} \citep{brown_verifiers_2025} API. We prefer verifiable rewards when tasks have objective checks (for example, strict answer-format validation), and fall back to a learned reward model for prompts without clean ground truth.

\section{Trinity Large Evaluation Results}

\subsection{Capability Benchmarks}
\label{sec:evals}

\begin{figure}
    \centering
    \includegraphics[width=1\linewidth]{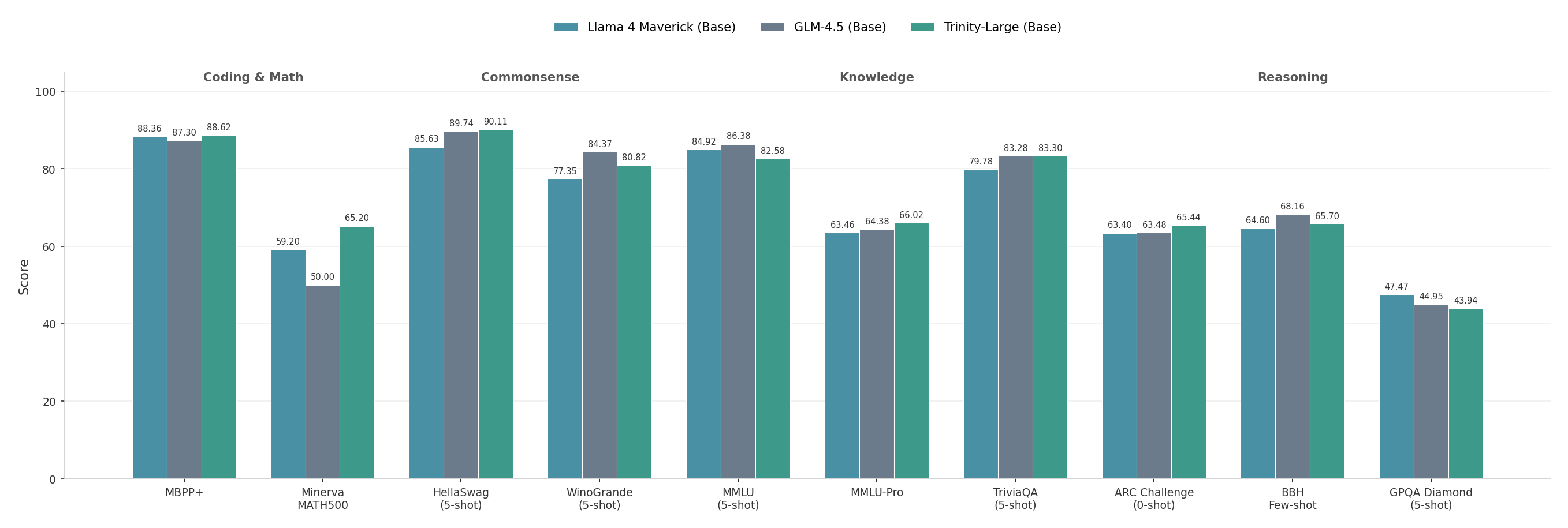}
    \caption{A comparison of Trinity Large Base to other similar open-weight base models.}
    \label{fig:base-evals}
\end{figure}

We evaluate Trinity Large Base on a standard benchmark suite spanning coding/math (MBPP+ \citep{liu2023codegeneratedchatgptreally}, Minerva MATH500 \citep{lewkowycz2022solvingquantitativereasoningproblems}), commonsense (HellaSwag \citep{zellers2019hellaswagmachinereallyfinish}, WinoGrande \citep{sakaguchi2019winograndeadversarialwinogradschema}), knowledge (MMLU \citep{hendrycks2021measuringmassivemultitasklanguage}, MMLU-Pro \citep{wang2024mmluprorobustchallengingmultitask}, TriviaQA \citep{joshi2017triviaqalargescaledistantly}, ARC Challenge \citep{clark2018thinksolvedquestionanswering}), and reasoning (BBH \citep{suzgun2022challengingbigbenchtaskschainofthought}, GPQA Diamond \citep{rein2023gpqagraduatelevelgoogleproofqa}). The evaluation results are reported in Table~\ref{tab:base-evals}, and comparisons to other open-weight base models are shown in Figure~\ref{fig:base-evals}. We note that Trinity Large Base achieves competitive scores with GLM 4.5 Base, despite having $4\times$ higher degree of sparsity and a roughly $2.5\times$ lower active parameter count.

\begin{table}[t]
\centering
\caption{Trinity Large Base performance on our base model evaluation suite.}
\label{tab:base-evals}
\begin{tabular}{l r}
\toprule
\textbf{Evaluation} & \textbf{Score} \\
\midrule
MBPP+ & 88.62 \\
Minerva MATH500 & 65.20 \\
HellaSwag (5-shot) & 90.11 \\
WinoGrande (5-shot) & 80.82 \\
MMLU (5-shot) & 82.58 \\
MMLU-Pro (5-shot) & 66.02 \\
TriviaQA (5-shot) & 83.30 \\
ARC Challenge (0-shot) & 65.44 \\
BBH (few-shot) & 65.70 \\
GPQA Diamond (5-shot) & 43.94 \\
\bottomrule
\end{tabular}
\end{table}

We also present evaluations for our instruct-tuned model, Trinity Large Preview, in Table \ref{tab:instruct-evals}. Our suite of instruct-tuned model benchmarks includes MMLU, MMLU-Pro, GPQA Diamond, SimpleQA \citep{wei2024measuringshortformfactualitylarge}, and AIME25.

\begin{table}[t]
\centering
\caption{Trinity Large Preview performance on our instruct model evaluation suite.}
\label{tab:instruct-evals}
\begin{tabular}{l r}
\toprule
\textbf{Evaluation} & \textbf{Score} \\
\midrule
MMLU & 87.21 \\
MMLU-Pro & 75.25 \\
GPQA Diamond & 63.32 \\
SimpleQA & 23.92 \\
AIME25 & 24.36 \\
\bottomrule
\end{tabular}
\end{table}

\subsection{Inference Benchmarks}
\begin{figure}
    \centering
    \includegraphics[width=\linewidth]{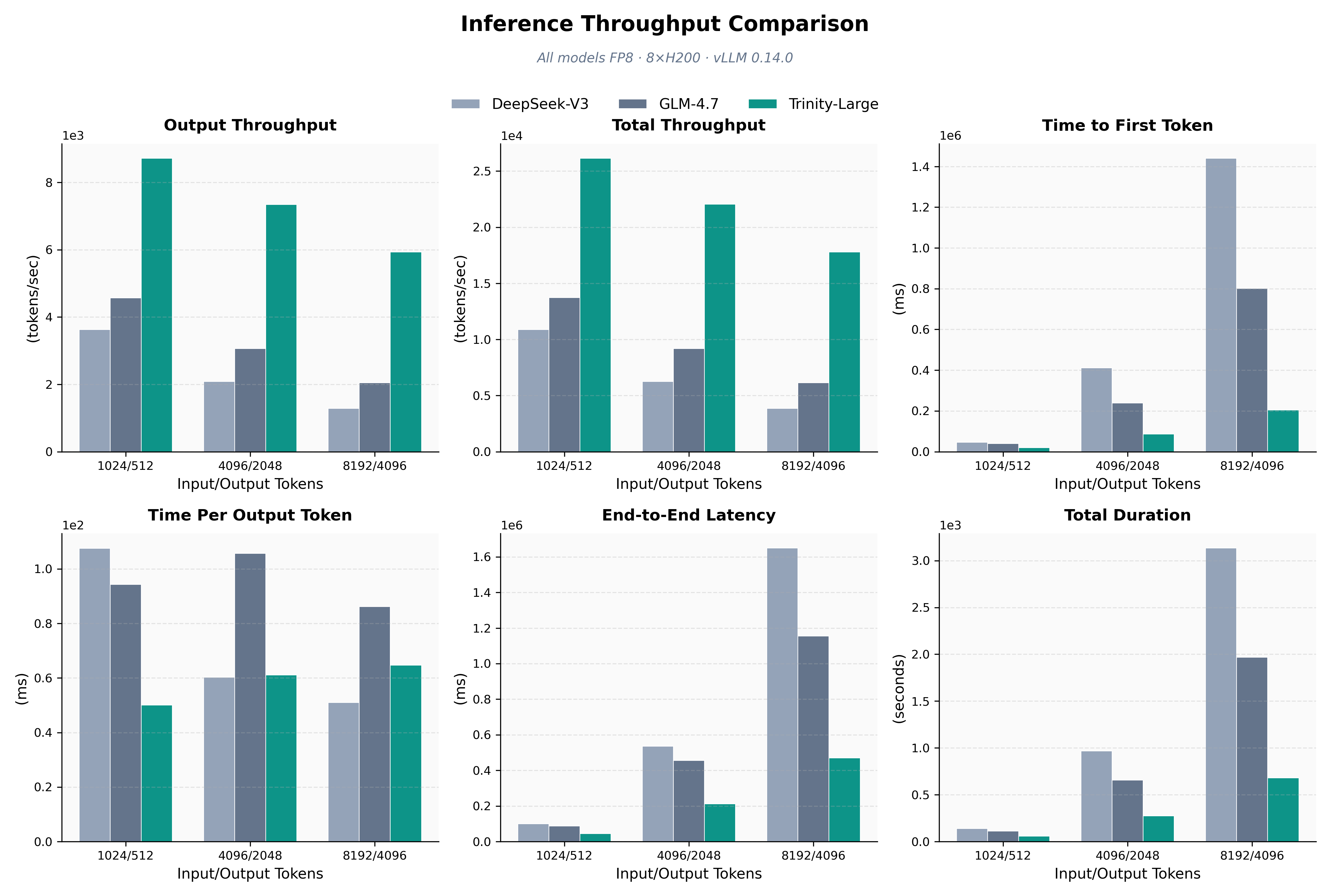}
    \caption{Throughput comparison of models. All tests were done with models quantized to FP8, using vLLM, on 8xH200.}
    \label{fig:inf}
\end{figure}

To benchmark inference performance, we use vLLM with all models quantized to FP8. Measurements were taken on an 8xH200 node. The measurements can be seen in Figure \ref{fig:inf}. The design of Trinity Large with its extreme sparsity and interleaved local/global attention results in strong performance.

\section{Discussion}

When training Trinity Large, we experienced some initial instability. On our first few runs, the loss decreased as expected and expert utilization looked roughly balanced early on. After some progress, routing behavior drifted and expert load became increasingly uneven, eventually resulting in collapsed experts. MaxVio (defined below) \citep{wang2024auxiliarylossfreeloadbalancingstrategy} would remain stable and then experience a sudden climb as the experts collapsed. When this happened, the loss plateaued, and evaluation improvements also plateaued, far from expected behavior. We attempted a few targeted changes: lowering the learning rate as a general method of stabilization, removing weight decay from normalization parameters, and increasing the post-norm $\epsilon$ value to reduce numerical sensitivity in normalization. Each of these changes improved behavior for a while, but did not prevent recurrence of the same failure mode. MaxVio continued to diverge and the loss trend did not recover to a stable downward trajectory. Since we had very limited time, we applied six changes at once, all targeted at increasing the stability of the run:
\begin{enumerate}
    \item We adopted our new load balancing method SMEBU, which we had briefly tested at very small scale.
    \item We disabled our MXFP8 kernels for linear layers and grouped GEMMs, falling back to BF16.
    \item We adopted z-loss with a small weight to stabilize training. \citep{wortsman2023smallscaleproxieslargescaletransformer, olmo20252olmo2furious}. Initially, we planned to use a weight of $1\times 10^{-4}$. However, due to a bug in the training code, the weight was being applied incorrectly, and was falling back to a z-loss weight of $0$. Out of concern due to the continued growth of the maximum logit value, we attempted to introduce z-loss in the middle of training (following \cite{marin_8b_retro}), and had to reduce the weight to $1\times 10^{-6}$, as larger values destabilized the network. This effectively stabilized the trend in maximum logit as well as mean logits.
    \item We swapped from a purely aux-loss-free load balancing strategy to including a sequence-wise aux loss with a small weight ($1\times 10^{-4}$), following \cite{deepseekai2025deepseekv3technicalreport}, \cite{5team2025glm45agenticreasoningcoding}, and \cite{coreteam2026mimov2flashtechnicalreport}.
    \item We chose to increase the number of dense layers from 3 to 6, to further stabilize representations.
    \item We chose to adopt intra-document masking to prevent token positions from attending to token positions from a different document, to reduce noise in the learning objective.
\end{enumerate}
With these fixes applied, MaxVio stopped diverging, expert utilization remained balanced throughout the run, and loss continued to smoothly converge to a lower value. Because the fixes were introduced together to unblock training, we did not have time to run controlled ablations to attribute stabilization to any individual change.

\noindent\textbf{MaxVio} \citep{wang2024auxiliarylossfreeloadbalancingstrategy} measures the relative worst-case expert overload, defined as
\begin{align}
\mathrm{MaxVio}
&=
\frac{\max_{i}\,\mathrm{Load}_{i}-\overline{\mathrm{Load}}}{\overline{\mathrm{Load}}}.
\tag{42}
\end{align}
where $\overline{\mathrm{Load}}$ is the mean expert load, and $\mathrm{Load}_{i}$ is the load on expert $i$ for $1 \leq i \leq N_r$.

\newpage
\section{Conclusion and Future Work}

In this report, we introduced the Trinity family of open-weight language models, culminating in Trinity Large, a model with 400B total parameters and 13B active per token. We also introduced Trinity Nano and Trinity Mini as smaller form factors and scaling-ladder validation points. At the largest scale, training stability and expert utilization balance were critical focus points.

We identify two promising directions for future work, both for us and the community at large. First, we believe that enabling greater sparsity in models overall will be the key to efficient scaling. Specifically in the case of MoEs, improved load balancing and routing will be important in allowing us to scale to greater sparsity while keeping training stable. Secondly, we view large-batch training as another key for efficient scaling. Algorithmic improvements that can push the critical batch size higher (and maintain sample efficiency and stability at scale) will directly translate into faster training and better utilization of modern hardware as the number of GPUs scales (when model parallelism is not needed).

\newpage
\section*{Additional Contributors}
\subsection*{Arcee AI}
Curt Larson\\
Scott Zembsch\\
Gabriel Santos\\
Ben Langer\\
Sam Fraser\\
Eric Lau\\
Mariam Jabara\\
James Weir\\
Davis Stone\\
Dante Simon\\
Molly Niland\\
Zachary Kirkendall\\
Mohit Khullar

\subsection*{Datology AI}
\subsubsection*{Technical Staff}
Amro Abbas\\
Rishabh Adiga\\
Cody Blakeney\\
Paul Burstein\\
Aldo Carranza\\
Spandan Das\\
Alvin Deng\\
Vineeth Dorna\\
Parth Doshi\\
Alex Fang\\
Tony Jiang\\
Siddharth Joshi\\
Brett Larsen\\
Jason Lee\\
Pratyush Maini\\
Kaleigh Mentzer\\
Luke Merrick\\
Haakon Mongstad\\
Ricardo Monti\\
Fan Pan\\
David Schwab\\
Darren Teh\\
Jason Telanoff\\
Jack Urbanek\\
Zhengping Wang\\
Josh Wills\\
Haoli Yin

\subsubsection*{Leadership}
Bogdan Gaza\\
Matthew Leavitt\\
Ari Morcos

\newpage
\bibliographystyle{plainnat}
\bibliography{references}

\end{document}